# Exploring difference in public perceptions on HPV vaccine between gender groups from Twitter using deep learning

Jingcheng Du

School of Biomedical Informatics

UTHealth

Houston, TX

jingcheng.du@uth.tmc.edu

Chongliang Luo

Department of Biostatistics, Epidemiology, and Informatics

UPenn

Philadelphia, PA

chongliang.luo@pennmedicine.upenn.edu

Qiang Wei

School of Biomedical Informatics

UTHealth

Houston, TX

qiang.wei@uth.tmc.edu

Yong Chen

Department of Biostatistics, Epidemiology, and Informatics

UPenn

Philadelphia, PA

ychen123@upenn.edu

Cui Tao†

School of Biomedical Informatics

UTHealth

Houston, TX

cui.tao@uth.tmc.edu

## ABSTRACT

In this study, we proposed a convolutional neural network model for gender prediction using English Twitter text as input. Ensemble of proposed model achieved an accuracy at 0.8237 on gender prediction and compared favorably with the state-of-the-art performance in a recent author profiling task. We further leveraged the trained models to predict the gender labels from an HPV vaccine related corpus and identified gender difference in public perceptions regarding HPV vaccine. The findings are largely consistent with previous survey-based studies.

## 1 Introduction

Vaccinations have led to dramatic declines in morbidity for various deadly vaccine-preventable diseases, such as polio and smallpox [5]. However, there was significant and increasing vaccine refusal and delay in the last two decades [17], which leaves the public unprotected and has been found associated with recent infectious disease outbreaks [29].

A good understanding of the public perceptions of vaccines is the first step towards developing effective vaccine promotion strategies[15,16]. Traditional survey methods suffer significant limitations (e.g. resource cost, inability in tracking changes) to understand the public perceptions [8]. The wide use of social media and the rise of advanced computational algorithms make social media a critical resource to understand public perceptions from a large population.

Demographic attributes, which are often collected by survey methods, are commonly missing in some social media platforms (e.g. Twitter). The lack of such information makes it challenging to investigate the difference in the public perceptions across different subpopulations, and complement the findings with traditional surveys.

Demographic attributes have been found associated with linguistic features of user-generated postings [26,28]. In this paper, we proposed a deep learning-based approach for gender prediction of Twitter users and then investigated gender difference on public perceptions regarding human papillomavirus vaccine (HPV) vaccine on Twitter. HPV infections cause about 33,700 cases of cancers every year in the U.S.[23]. Although, HPV vaccines are effective to prevent HPV induced cancer, the vaccine rates remain to be suboptimal. We target HPV vaccine as the use case and the major contributions of this paper are three-folds:

1. We proposed a convolutional neural network model with embedding fusion for gender prediction using English Twitter text as input, which compared favorably with the state-of-the-art performance in a recent author profiling task.
2. We conducted a comparison on machine-learning and deep-learning algorithms in Twitter gender prediction.
3. We evaluated an HPV vaccine related Twitter corpus and identified the gender difference in public perceptions regarding HPV vaccine. The findings are largely consistent with previous survey-based studies.

## 2 Related work

Recent efforts have developed machine learning and deep learning-based approaches to understand the public perceptions on vaccines from social media [2,9]. However, due the lack of availabilities of gender attributes in Twitter users, few studies investigated the gender difference in public perceptions in vaccines. Huang et al leveraged the Demographer [14] to infer gender information from Twitter users [12] and studied the how vaccine tweets counts vary across genders. However, Demographer and other name-based inferring tools [30,31] suffer limitations when the name information of Twitter users is not available or accurate.

Previous efforts framed gender prediction as binary classification tasks and proposed machine learning-based approaches (e.g. support vector machines) with extensive feature engineering from Twitter text. Word and character-level n-grams based-features were mostly commonly used [1,3]. Additional features including Emojis, part-of-speech (POS) tags, latent semantic analysis(LSA), lexicon features were also adopted in many systems [7,22]. Traditional machine learning-based approaches achieved high performance on Twitter gender prediction challenges.

Deep learning-based methods can save great efforts in feature engineering and have achieved state-of-the-art performance in many natural language processing (NLP) tasks[27]. Previous efforts have investigated the use of deep learning in gender prediction. However, compared with traditional machine learning base approaches with feature engineering, their performances using Twitter textual data were suboptimal. For example, in 6th Author Profiling Task at PAN 2018 [20], the top three systems in gender prediction (using English Twitter text only) all adopted non-deep learning-based systems. Sierra1 et al leveraged a feed-forward neural network with fasttext embedding and ranked 4th in the English Twitter text category[24]; Takahashi et al designed a recurrent neural network for text and ranked 7th in that category[25]. In 5th Author Profiling Task at PAN 2017[21], Miura et al proposed a neural attention network to integrate word and character information[18], however, only ranked 6th in English gender prediction.

## 2  Method

There are two major steps in this study: 1) to evaluate a convolutional neural network-based deep learning model for English Twitter gender prediction and evaluate the model on the most recent open challenge task: 6th Author Profiling Task at PAN 2018 [20]; 2) to leverage the trained model on the gender prediction of Twitter users who have discussed HPV vaccine related topics and investigate the gender difference on the public perceptions regarding HPV vaccines.

### 2.1  Datasets

Author profiling tasks at PAN is a series of international challenges which aim to classify the texts into classes based on the stylistic choices of their authors. Author profiling tasks at PAN 2018 focused on gender identification in Twitter. For each Twitter user, a total of 100 tweets and 10 images were provided and the users were grouped by three language including English, Arabic and Spanish. In this study, we focused on English tweets with text data only (image data were excluded). The task provided a balanced corpus with regard to gender, which contained 3,000 Twitter users for training and 1,900 Twitter users for testing[20]. We used the training data set to develop our gender prediction model and evaluated the performance on the testing data set.

We used a set of HPV vaccine related keywords to collect 1,489,272 English tweets by using Twitter streaming API from Jan. 1, 2014, to Oct. 26, 2018. We annotated a subset of 6,000 tweets based on their relevance to four Health Belief Model (HBM) constructs (i.e. perceived susceptibility, perceived severity, perceived benefits, and perceived barriers) [9] and one Theory of Planned Behavior (TPB) construct (i.e. attitude)[10]. The gold standard data was used to train and evaluate the attentive recurrent neural network described in [9]. We repeated random sampling of the tweets and the training 10 times for each construct and the final prediction of the all the un-labeled tweets was based on the community ensemble (i.e. majority voting) of the 10 models. We then collected up to 100 recent tweets from each Twitter users in our corpus. We removed the tweets of which the Twitter accounts are not valid during collection period (Dec. 2018 – Jan. 2019), 1,090,208 tweets were included in our final analysis. 774,112 tweets were classified as related to HBM, among which 78,151, 183,051, 216,950 and 321,164 tweets were classified as perceived susceptibility, perceived severity, perceived benefits, and perceived barriers respectively. 785,287 tweets were classified as related to TPB attitude, among which 290,129, 284,550 and 210,608 tweets were classified as positive, negative, and neutral respectively.

### 2.2  Convolutional neural network (CNN) with embedding fusion

The overall architecture of proposed framework is illustrated in Figure 1. Specifically, we design a character layer, which takes the character embedding of each character of the tweet token and outputs the summary of characters for each token using a convolutional layer followed by a max-pooling layer. The character layer ensures mapping both the in-vocabulary words and the out-of-vocabulary words (e.g. incorrect spellings) to a high dimensional vector. The word layer concatenates the output of the character layer, word embedding and POS embedding together to comprehensively capture linguistic features of each word. The output of the word layer is fed to another convolutional layer and max-pooling layer to represent the information of the Twitter user. We add a dense layer with batch normalization on top of pooling layer. The output layer is a fully connected layer with Softmax outputs. We further add L2 regularization and dropout to avoid overfitting. The major parameter setting for the proposed model can be seen in Table 1. We term this architecture CNN_char _pos.

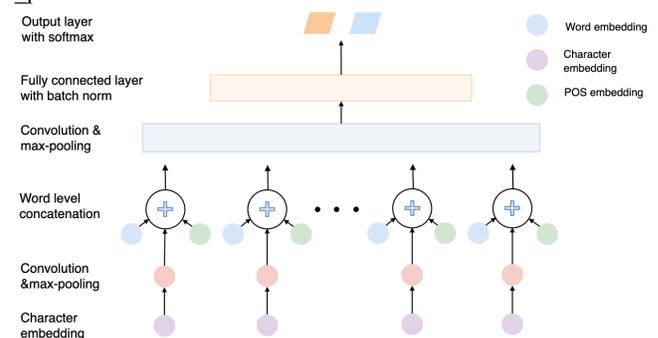

**Figure 1 The architecture of proposed convolutional neural network**

### 2.3  Experiment setting

*2.3.1 Data preprocessing.* For each tweet, we first followed [19] for pre-processing (e.g. user name normalization, URL normalization, lowercase), then leveraged NLTK 3.4.1 TweetTokenizer for tokenization and Taggers for Part of Speech (POS) Tagging. For each user, we combined all the 100 tweets into a single document, which was served as the input for the machine learning and deep learning models.

*2.3.2 Baseline models.* We chose competitive machine learning and deep learning models for comparison. For traditional machine learning algorithms, we used *term frequency-inverse document frequency* (TF-IDF) as features and evaluated several algorithms including support vector machine (SVM), logistic regression (LR), and extra tress (ET). For deep learning algorithms, we evaluated the basic CNN model with word embedding only (we term it CNN), CNN model with character embedding (we term it CNN_char) and an attentive bi-directional recurrent neural network model (we term it RNN).

*2.3.3 Cross fold validation & community ensemble.* We split the training data set into 5 folds. For each fold, 4 folds data were used as training corpus while the last fold was used as validation corpus. The model which achieved the highest accuracy among all epochs on the validation corpus was selected and evaluated on the official testing data. We calculated the mean accuracy of 5 folds on the testing data and the accuracy of ensemble models from 5 folds after majority voting.

Table 1 Parameters setting for the proposed neural network

| Parameter | Setting |
|---|---|
| Learning rate | 0.001 |
| Batch size | 64 |
| POS embedding dimension | 10 |
| Character embedding dimension | 50 |
| Pre-trained word embedding | GloVe Twitter embedding (d=200) |
| L2 regularization | 0.00001 |
| Dropout rate | 0.2 |
| No. of filters: word/character level | 2048/50 |
| Filter size: word/character level | 1,2,3/3 |

## 2.4 Chi-square test

We then investigated the difference in the perceptions of HPV vaccine between genders through Chi-square test. The difference is measured by the odds ratio of gender vs the perceptions with respect to numbers of tweets mapped to constructs of HBM and TPB for each year, using twitter data from 2014-2018. The year 2018 has twitter data only up to October. Chi-square tests are used to determine whether there is a significant difference between the frequencies between the two gender groups. To adjust for multiple comparison, Bonferroni correction is used with nominal significance level 0.05 and number of comparisons 25 (5 tests each year).

## 3 Results and discussion

## 3.1 Comparison of models in gender prediction

The mean, standard deviation (SD) and voting accuracy of 5 folds for each algorithm on the testing dataset can be seen in Table 2. CNN based models outperformed SVM and RNN models on the Twitter gender prediction tasks. The ensemble of CNN_char_pos achieved the highest accuracy among all models and higher accuracy compared to the best results (0.8221 from [7]) reported in Task at PAN 2018. The ensemble model improved the accuracy for all the algorithms

Table 2 Comparison of algorithms on Twitter gender prediction

|  | SVM | RNN | CNN | CNN_char | CNN_char_pos |
|---|---|---|---|---|---|
| Mean | 0.7902 | 0.7874 | 0.8019 | 0.8127 | 0.8128 |
| SD | 0.0035 | 0.0106 | 0.0066 | 0.0018 | 0.0060 |
| Voting | 0.7968 | 0.8047 | 0.8153 | 0.8189 | 0.8237 |

## 3.2 Gender difference in public perceptions

The trained ensemble of CNN_char_pos was then used to infer the user gender in our HPV Twitter corpus. Out of 1,090,208 tweets in the corpus, 572,314 tweets (52.50%) were inferred to be sent by female Twitter users. We further evaluate the coverage of the ensemble model by calculating the average of prediction probability (i.e. the value of Softmax output) of 5 folds on the prediction of HPV Twitter corpus. 818,908 tweets (75.11%) in the corpus have higher average probability than 0.80, which shows the high coverage of our model.

The odds ratios of gender vs HBM and TPB measures along year 2014-2018 are plotted in Figure 2. It is shown that for all the constructs, male have lower positive rate than female, except for HBM barriers, which the odds ratio is above 1.7 for all the 5 years. The Chi-square tests for all the odds ratios are significant, with p-value less than 0.001. The results are largely consistent with the findings from previous survey-based studies. For example, [13] found that men perceived more barriers to HPV prevention than did women, while women perceived more benefits and knowledge to HPV vaccine in a Korean population; [4] found men scored higher on the perceived barriers to HPV vaccine, while lower on perceived severity, perceived benefit than did women in a population of African-American college students. One limitation of this study is that we treated predicted gender as true gender in following Chi-square test, which could lead to information bias due to the misclassification rates [6,11]. Besides, Twitter population is not well representative to the general population, which could lead to additional bias to our analysis.

## 4 Conclusion

In this study, we proposed a CNN model with embedding fusion for Twitter gender inference using English tweet text as input. The performance of proposed model compared favorably with the state-of-the-art performance in a recent author profiling task. The comparison of embedding fusion shows the efficacy of using character embedding and POS embedding in Twitter gender prediction. We leveraged the trained models on an HPV vaccine related Twitter corpus and identified the public perception difference regarding HPV vaccine between gender groups, which are largely consistent with previous survey-based studies. This study shows the potentials of using social media and deep learning models to understand differences in the public perceptions of public health related topics for different demographic groups.

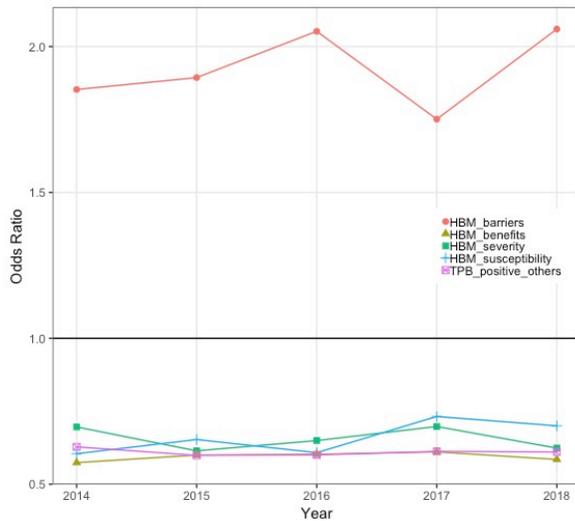

**Figure 2 Difference in frequencies of tweets aligned to HBM and TPB constructs on HPV vaccine between genders. Odds ratios along year 2014-2018 are presented. Year 2018 has partial data up to October. Odds ratio greater than 1 means male have higher positive rate than female.**

## ACKNOWLEDGMENTS

This research was supported by the National Institutes of Health under Award Number 2R01LM010681–05 and R01 LM011829 and the Cancer Prevention Research Institute of Texas (CPRIT) Training Grant #RP160015.